\documentclass{article}
\usepackage{spconf,amsmath,graphicx}

\usepackage{lipsum}
\usepackage{bbm}
\usepackage{amssymb}
\usepackage{xcolor}
\usepackage{booktabs}
\usepackage{pifont}
\usepackage{diagbox}
\usepackage{fontawesome}
\usepackage[bookmarks=false,hidelinks]{hyperref}
\usepackage{textcomp}
\usepackage[T1]{fontenc}

\newcommand{\inR}{\in \mathrm{R}}

\definecolor{green}{HTML}{009900}
\definecolor{orange}{HTML}{FF8000}
\definecolor{blue}{HTML}{6C8EBF}
\definecolor{dirtyyellow}{HTML}{D6B656}
\definecolor{new}{HTML}{6C8EBF}

\title{SYNCHFORMER: EFFICIENT SYNCHRONIZATION FROM SPARSE CUES}

\name{Vladimir Iashin$^{1,3}$\thanks{This research was funded by the Academy of Finland projects 327910 and 324346, EPSRC Programme Grant VisualAI EP$\slash$T028572$\slash$1, and a Royal Society Research Professorship.  We also acknowledge CSC (Finland) for awarding this project access to the LUMI supercomputer, owned by the EuroHPC JU, hosted by CSC and the LUMI consortium through CSC.} \qquad Weidi Xie$^{2,3}$ \qquad Esa Rahtu$^{1}$ \qquad Andrew Zisserman$^{3}$}
\address{$^{1}$Tampere University \qquad $^{2}$Shanghai Jiao Tong University \qquad $^{3}$University of Oxford}

\begin{document}
\maketitle

\begin{abstract}
Our objective is audio-visual synchronization with a focus on `in-the-wild' videos,
such as those on YouTube, where synchronization cues can be \textit{sparse}.
Our contributions include a novel audio-visual synchronization model, and training that decouples feature
extraction from synchronization modelling through multi-modal segment-level contrastive pre-training.
This approach achieves state-of-the-art performance in both \textit{dense} and \textit{sparse} settings.
We also extend synchronization model training to AudioSet a million-scale `in-the-wild' dataset,
investigate evidence attribution techniques for interpretability,
and explore a new capability for synchronization models: audio-visual synchronizability.
\href{https://www.robots.ox.ac.uk/~vgg/research/synchformer/}{\color{new} \textbf{\texttt{robots.ox.ac.uk/\textasciitilde vgg/research/synchformer}}}

\end{abstract}
%
\section{Introduction}
\label{sec:intro}

The task of audio-visual synchronization is to predict the temporal offset between audio and visual streams in a video. 
Previous work in this area has mostly addressed the scenario where \textit{dense} cues are available, such as lip movements in talking heads or instrument
playing~\cite{Chung2019perfect,Chung16a,Afouras20b,arandjelovic2018objects,Owens18audio-visual}. While for open-domain videos, e.g.\ YouTube videos of dogs barking or isolated actions, the synchronization cues are {\em sparse}, and only available at certain time intervals~\cite{chen2021audio}. Naturally, this brings challenges 
in that an extended temporal window is required so that sparse cues are not missed.

\begin{figure*}[tb]
    \centering
    \resizebox{0.9\textwidth}{!}{
        \includegraphics{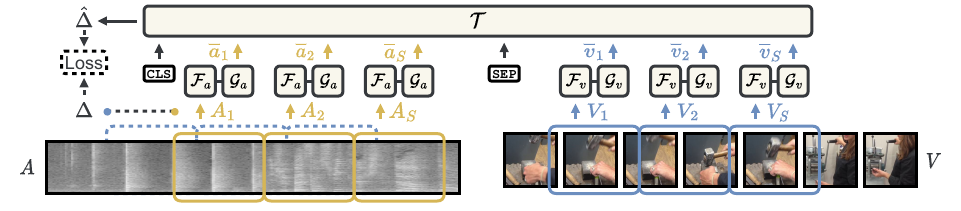}
    }
    \caption{
      \textbf{Synchformer} ($\mathcal{M}$).
      The audio and visual streams ($A, V$) are split into $S$ segments of equal duration.
      Then, the segment-level inputs are fed into their respective feature extractors
      ($\mathcal{F}_a, \mathcal{F}_v$).
      The streams are aggregated along the space (or frequency) by
      $\mathcal{G}_a, \mathcal{G}_v$, and concatenated into a single sequence with auxiliary tokens
      (\texttt{CLS} and \texttt{SEP}).
      The sequence is fed into the synchronization module $\mathcal{T}$, which predicts
      the temporal offset $\hat{\Delta}$.
      The dashed lines show the training of the model.
    }
    \label{fig:method}
\end{figure*}

The synchronization of \textit{sparse} signals was explored in~\cite{sparse2022iashin}. However, this approach has several limitations, 
first, it relies on the availability of a dataset with \textit{dense} synchronization cues for pre-training, second, it requires end-to-end training which, in turn, poses a high memory footprint and limits the selection of feature extractors.
In this work, we address these limitations by training the model in two stages: first, segment-level contrastive pre-training of feature extractors, 
and second, training of a light-weight synchronization module.
In addition, we explore evidence attribution techniques for interpretability and extend the capability of the model to synchronizability prediction, i.e. if it is possible to synchronize the provided audio and visual streams.
Finally, we scale up the training to AudioSet~\cite{gemmeke2017audio}, 
a million-scale `in-the-wild' dataset, and achieve state-of-the-art results in both \textit{dense} and \textit{sparse} settings.

\section{Related work}
Early works primarily focused on synchronizing the videos of human presentations, for example,
\cite{hershey1999audio,slaney2000facesync} used hand-crafted features and statistical models. In the deep learning era, \cite{chung2016lip} introduced a self-supervised two-stream architecture trained using a contrastive loss.
Subsequent enhancements included multi-way contrastive training~\cite{Chung2019perfect} and the incorporation of Dynamic Time Warping~\cite{rabiner1993fundamentals, halperin2019dynamic}. \cite{Khosravan19} showcased the advantages of spatio-temporal attention, while~\cite{kim2021end} introduced a cross-modal embedding matrix to predict synchronization offsets.
Building on this progress, \cite{kadandale2022vocalist} presented an architecture featuring a set of transformers for cross-modal feature extraction and fusion, while \cite{gupta2023modeformer} explored contrastive pre-training to detect if a speech video is out-of-sync.

Research in the synchronization of diverse classes of videos was sparked by
Chen et al.\ \cite{chen2021audio} who employed a transformer and leveraged
a subset of VGGSound \cite{Chen20a}, addressing 160 distinct classes.
Iashin et al.\ \cite{sparse2022iashin} explored synchronization of videos with \textit{sparse} temporal cues and proposed to use learnable query vectors to pick useful features from the audio and visual streams to reduce the computation burden, however, the approach required a dataset with \textit{dense} synchronization cues for pre-training, and the model can be trained only end-to-end which poses a high computation burden that
limits the selection of feature extractors.
                    
\section{Method}
\label{sec:method}

Given audio and visual streams $A$ and $V$, a synchronization model $\mathcal{M}$ predicts the temporal offset $\Delta$ between them, i.e.\ $\Delta = \mathcal{M}(A, V)$. Fig.~\ref{fig:method} shows an overview of our
 {\em Synchformer} ($\mathcal{M}$) synchronization model.
Instead of extracting features from the entire video, we extract features from shorter temporal segments (0.64 sec) of the video.
The segment-level audio and visual inputs are fed into their respective feature extractors independently to obtain frequency and spatio-temporal 
features.
Then, the streams are aggregated along the space (or frequency) dimensions in $\mathcal{G}_a, \mathcal{G}_v$. The synchronization module $\mathcal{T}$ inputs the concatenated sequence of audio and visual aggregated features and predicts the temporal offset $\hat{\Delta}$.

\subsection{Architecture of Synchformer ($\mathcal{M}$)}
\label{sec:architecture}

\paragraph*{Segment-level features.}
A video with audio (a mel spectrogram) and visual streams ($A, V$) is split into $S$
segments of equal duration, $A_s \inR^{F \times T_a}$  ($F, T_a$ are frequency and time dimensions), and $V_s \inR^{T_v \times H \times W \times 3}$ ($T_v, H, W, 3$ are time, height, width, and RGB).
For each segment $s$, we obtain audio and visual feature maps, $a_s \inR^{f \times t_a \times d}, v_s \inR^{t_v \times h \times w \times d}$ as
\begin{align}
  a_s = \mathcal{F}_a(A_s) \qquad v_s = \mathcal{F}_v(V_s) \qquad s \in \{1, \dots, S\},
  \label{eq:feature_extraction}
\end{align}
where $\mathcal{F}_a$ is AST \cite{gong2021ast} and $\mathcal{F}_v$ is Motionformer with divided space-time attention 
\cite{patrick2021keeping,bertasius2021space},
$a_s$ and $v_s$ are outputs of last layers of $\mathcal{F}_a$ and $\mathcal{F}_v$.
To reduce the sequence length, we aggregate the frequency ($f$) and spatial ($h\times w$) dimensions as,
\begin{align}
  \overline{a}_s = \mathcal{G}_a\big([\text{\texttt{AGG}}, a_s]) \qquad \overline{v}_s = \mathcal{G}_v([\text{\texttt{AGG}}, v_s]\big)
  \label{eq:agg}
\end{align}
where $\mathcal{G}_{a/v}$ are single-layer transformer encoders~\cite{vaswani2017attention,devlin2018bert}, $\overline{a}_s \inR^{t_a \times d}$ and $\overline{v}_s \inR^{t_v \times d}$, 
$[,]$ is a concatenation operator,
and \texttt{AGG} is learnable token whose output is used as the aggregation $\overline{a}_s, \overline{v}_s$.
Trainable positional encodings are added.

\paragraph*{Synchronization module.}
The audio and visual features $\overline{a}_s, \overline{v}_s$ from all segments $S$ are concatenated with learnable tokens, along the time dimension, and then fed into the synchronization module $\mathcal{T}$
that predicts the temporal offset $\hat{\Delta}$,
\begin{align}
  \hat{\Delta} = \mathcal{T}\big([\texttt{CLS}, \overline{a}_1, \dots, \overline{a}_S, \texttt{SEP}, \overline{v}_1, \dots, \overline{v}_S]\big),
  \label{eq:sync}
\end{align}
where $\texttt{SEP}$ is a separating token, and $\mathcal{T}$ is a transformer encoder with $3$ layers,
$8$ heads, and $d=768$.
Following \cite{sparse2022iashin}, we pose synchronization as a classification problem with a fixed number of classes.
We add trainable positional encoding to the input sequence.
To predict $\hat{\Delta}$, we apply a linear layer with softmax activation to the output of the \texttt{CLS} token.

\begin{figure}[tb]
    \raggedright
    \resizebox{0.50\textwidth}{!}{
        \includegraphics{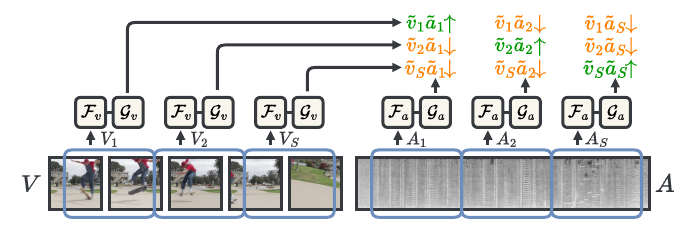}
    }
    \caption{
      \textbf{Segment AVCLIP Pre-training.}
      The audio ($A$) and visual ($V$) streams are split into $S$ segments, which are
      fed into their respective feature extractors ($\mathcal{F}_a, \mathcal{F}_v$).
      The outputs of the feature extractors ($\overline{a}_s, \overline{v}_s$) are aggregated along time (omitted for clarity)
       to obtain audio and visual features ($\tilde{a}_s, \tilde{v}_s$).
      The features from corresponding segments in a video are pulled together ({\color{green} $\uparrow$}),
      while the features from other segments are pushed apart ({\color{orange} $\downarrow$}).
    }
    \label{fig:s1_training}
\end{figure}

\subsection{Training}
\label{sec:training}

The model is trained in two stages.
First, audio and visual feature extractors (Eq.~\eqref{eq:feature_extraction}, \eqref{eq:agg}) undergo pre-training with segment-level contrastive learning, facilitating the extraction of distinguishable features from each segment in a video.
Second, the synchronization module (Eq.~\eqref{eq:sync}) is trained to predict the temporal offset between audio and visual streams, using features from 
 the pre-trained and frozen feature extractors (whose weights are not updated during this stage).

\paragraph*{Segment-level audio-visual contrastive pre-training.}

Drawing on the success of CLIP \cite{radford2021learning},
we train the model with InfoNCE loss~\cite{oord2018representation} to distinguish between positive and negative pairs.
In our setting, a positive pair is a pair of audio and visual segments from the same time interval of the same video, 
while negative pairs consist of segments from the same video and segments from other videos in the batch.

To obtain logits for each segment, we average $\overline{a}_s, \overline{v}_s$ along the time dimension
($t_a, t_v$), project them with a linear layer, and normalize to unit length,
yielding $\tilde{a}_s, \tilde{v}_s$.
Thus, for a batch of $B$ audio-visual pairs (videos) with $S$ segments in each,
the segment-level contrastive loss is defined by
\begin{align}
  \mathcal{L}^\text{I}_{a\rightarrow v} = -\frac{1}{BS} \sum_{i=1}^{BS} \log \frac{\exp(\tilde{a}_{i} \cdot \tilde{v}_i / \tau)}
  {\sum_{j=1}^{BS} \exp(\tilde{a}_i \cdot \tilde{v}_j / \tau)},
\end{align}
where $\tau$ is a trainable temperature parameter.
The counterpart loss ($\mathcal{L}^\text{I}_{v\rightarrow a}$) is defined analogously.
The total contrastive loss is obtained by averaging the two,
$\mathcal{L}^\text{I} = (\mathcal{L}^\text{I}_{a\rightarrow v} + \mathcal{L}^\text{I}_{v\rightarrow a}) / 2$.
With $B=2$ videos and $S=14$ segments per video (8.96 sec, in total), the
effective batch size is $BS=28$.
In our experiments, higher $B$ or using the momentum queue did not translate
to better synchronization performance.
We refer to this as \textit{Segment AVCLIP} and outline it in Fig.~\ref{fig:s1_training}.

\paragraph*{Audio-Visual synchronization module training.}
We assume that the audio and visual streams in the training data are synchronized. We formulate synchronization as a classification problem with a fixed number of classes, following~\cite{sparse2022iashin}.
During the training of this stage, we rely on the audio and visual features $\overline{a}_s, \overline{v}_s$ obtained
\textit{before} the time-wise within-segment aggregation, as in Eq.~\eqref{eq:agg}.
Segments are extracted with a temporal step size that overlaps half the segment length, as this was found to improve synchronization performance.
We use a batch size of 16 videos with 14 segments (4.8 sec, in total) in each.
The model is trained with cross entropy loss.

\subsection{Discussion}
Our approach yields several advantages over prior designs~\cite{sparse2022iashin}:
(i) Since the feature extractors (transformers) now operate on shorter input sequences and the synchronization module is trained later, we can allocate more trainable parameters and get higher-quality features;
(ii) Provided that the features from different segments in a video are distinguishable,
a light-weight synchronization module can be trained while the weights of feature extractors
are \textit{frozen} (not updated);
(iii) It streamlines the temporal evidence attribution visualization for the synchronization task;
(iv) It is adaptable to new downstream tasks, such as synchronizability prediction.

\section{Additional Capabilities}
\label{sec:capabilities}

\subsection{Evidence Attribution}
\label{sec:evidence}
We aim to determine the temporal evidence used by the model for synchronization predictions, 
assigning an attribution score to each temporal interval of the audio and visual streams.
We assume that if a part of the input is not important, the model can predict the offset correctly without it.

To determine the importance of a temporal interval, we randomly sample a mask $M_k \inR^T\in[0, 1]$ (if $M_{k,t}=0$ the content is masked), where $T=S(t_v+t_a)$, and apply it to the input of the synchronization module $\mathcal{T}$.
For each randomly masked input, we observe the model's prediction and, if the prediction is correct ($\pm$0.2 sec), we record the mask.
To get a reliable score, we repeat this process multiple times ($K$).

The attribution score of a temporal interval $t$ is defined as $\sum_{k=1}^{K} \mathbbm{1}(M_{k,t} = 1 \land \hat{\Delta}_i = \Delta_i)/K$.
The score is closer to 1 if the interval is important, and closer to 0.5 if it is less important.
Considering a large number of potential permutations, we mask chunks of audio and visual `frames' at once, i.e.\ 0.1 sec, and operate on one modality at a time. 
As a masking strategy, we replace the content in selected intervals with the content from a random `distractor' video from the test set.

\subsection{Predicting Synchronizability}
\label{sec:predicting_syncronizability}
In contrast to the general assumption in existing work, that all videos are synchronizable, we additionally explore the possibility of inferring the synchronizability of provided audio and visual streams, i.e.\ if it is possible to synchronize them at all.
To train a model for this task, we fine-tune \textit{Synchformer} by adding another
binary synchronizability prediction head.
We use the same training data as for the synchronization training and uniform sampling of the offset target class, but with 0.5 probability use an offset that is equal to the duration of the input track to ensure that the streams are not synchronizable.
The model is trained with binary cross entropy loss.

\section{Experiments}
\label{sec:experiments}

\paragraph*{Datasets.}

For the \textit{dense} setting, we use $\sim$58k clips from the LRS3 dataset~\cite{afouras2018lrs3} processed as in~\cite{sparse2022iashin},
i.e.\ the videos are not cropped to the face region (`full scene').
For the \textit{sparse} setting, we use the VGGSound~\cite{Chen20a} dataset that contains
$\sim$200k `in-the-wild` 10-second YouTube videos
($\sim$186k after filtering for missings).
We train both stages of our model on this dataset (§\ref{sec:training}).
For evaluation, we use VGGSound-Sparse~\cite{sparse2022iashin}, a curated subset of
12 \textit{sparse} data classes from
VGGSound-Sparse~\cite{sparse2022iashin} which consists of $\sim$600 videos
(542 after filtering)
and VGGSound-Sparse (Clean) where all input clips (439 videos) have one or more synchronizable events.
In addition, we also use the AudioSet~\cite{gemmeke2017audio} dataset which consists of $\sim$2M ($\sim$1.6M after filtering) `in-the-wild' YouTube videos.

\paragraph*{Metrics.}
As a measure of synchronization performance,
we use the accuracy of the top-1 prediction of the temporal offset across $21$ classes from $-2$ to $+2$ seconds, with $0.2$ seconds granularity.
Similar to \cite{sparse2022iashin}, we allow for $\pm0.2$ seconds tolerance ($\pm1$ class) for the prediction to be considered correct.
For synchronizability prediction, we use binary accuracy.

\paragraph*{Baseline.}
We compare our approach to the state-of-the-art method for audio-visual synchronization,
SparseSync~\cite{sparse2022iashin}, which relies on ResNet-18~\cite{he2016deep} and S3D~\cite{xie2018rethinking}
as audio and visual feature extractors, and two DETR-like~\cite{carion2020end}
transformers to pick useful features from the audio and visual streams, and finally, the refined features
are passed to the transformer encoder to predict the temporal offset.

\subsection{Synchronization Results}

\begin{table}[tb]
  \raggedright
  \footnotesize
  \setlength{\tabcolsep}{3pt}
    \begin{tabular}{l l r r c}
      \multicolumn{5}{l}{\textbf{Dense $\downarrow$}} \\
      \toprule
        & \textbf{Train}  & \textbf{Params}  & \textbf{Train \faClockO }     & \textbf{LRS3 (`Full Scene')}    \\
      \textbf{Method} & \textbf{dataset} & \textbf{($\times 10^6$)} & \textbf{($\times10^3$)} & \textbf{A@1} / \textbf{$\pm$1 cls}    \\
      \midrule
      AVST \cite{chen2021audio,sparse2022iashin} & LRS3-FS               & 32.3            & 0.2 & 58.6 / 85.3 \\
      SparseS \cite{sparse2022iashin}         & LRS3-FS               & 55.3            & 0.9 & 80.8 / 96.9 \\
      Synchformer                                & LRS3-FS               & 214+22.6      & 0.4+0.6 & \textbf{86.5 / 99.6} \\
      \bottomrule
    \end{tabular}
    \\
  \setlength{\tabcolsep}{2pt}
  \begin{tabular}{l l r r c c c}
    \multicolumn{6}{l}{\textbf{Sparse $\downarrow$}} \\
    \toprule
      &   \textbf{Train} & \textbf{Params} & \textbf{Train \faClockO} & \textbf{VGS-Sp} & \textbf{VGS-Sp (C)}        \\
    \textbf{Method}                & \textbf{dataset}          & \textbf{($\times 10^6$)} & \textbf{($\times10^3$)} & \textbf{A@1 / $\pm$1 cls} & \textbf{A@1 / $\pm$1 cls}\\
    \midrule
      AVST \cite{chen2021audio,sparse2022iashin} & VGS-Sp$^*$ & 32.3        & 0.3 & 14.1 / 29.3                 & 16.6 / 32.1                \\
      SparseS \cite{sparse2022iashin}         & VGS-Sp$^*$ & 55.3        & 1.0 & 26.7 / 44.3                 & 32.3 / 50.0                \\
      SparseS \cite{sparse2022iashin}         & VGS$^*$        & 55.3        & 1.8  & 33.5 / 51.2                 & 43.4 / 62.1                \\
      SparseS \cite{sparse2022iashin}$^\dagger$     & AudioSet$^*$   & 55.3        & 19.7 & 35.3 / 56.7                 & 40.0 / 63.0                \\
      Synchformer                                & VGS            & 214+22.6 & 0.5+2.1 & \textbf{43.7 / 60.2}        & \textbf{52.9 / 70.1}                \\
      Synchformer                                & AudioSet       & 214+22.6 & 4.2+4.4 & \textbf{46.8 / 67.1}        & \textbf{54.6 / 77.6}          \\
    \bottomrule
  \end{tabular}
  \caption{
    \textbf{Synchronization results for \textit{dense} (top) and \textit{sparse} signals.}
    The \textit{dense} setting results are reported on the test set of LRS3-FS (`Full Scene'), and for the \textit{sparse} setting, the test set of VGGSound-Sparse (VGS-Sp) and VGS-Sp Clean (C) are used.
    The top-1 accuracy is reported across 21 offset classes without and with $\pm$1 class tolerance (indicated by the value before and after `/').
    The train time is shown in GPU hours.
    For \textit{Synchformer}, we show the combined (+) number of parameters and training time for both stages.
    $^\dagger$:~trained by us; $^*$: pre-trained on LRS3 (`Full Scene').
  }
  \label{tab:sync_results_dense_sparse}
\end{table}

In Tab.~\ref{tab:sync_results_dense_sparse} (top), we report the results on the \textit{dense} setting, i.e.\ by training and testing on the LRS3 (`Full Scene') dataset.
According to the results, our model outperforms the state-of-the-art method by a significant margin.

The results on the \textit{sparse} setting are shown in Tab.~\ref{tab:sync_results_dense_sparse} (bottom), i.e.\
by training VGGSound and AudioSet and evaluating on the test set VGGSound-Sparse or a clean version of it.
Even though the baseline methods were pre-trained on \textit{dense} data (LRS3 (`Full Scene')),
our model outperforms them by a significant margin after training only on \textit{sparse} data, i.e.
rows 3 vs.\,5 and 4 vs.\,6,
which showcases the benefits of the proposed approach.
Notice that, with the two-stage approach, 
we can train much larger models (55.3M vs 200+M), 
which results in better performance, yet the training time is comparable on smaller datasets (VGGSound and VGGSound-Sparse) and much less on the larger one (AudioSet).

In addition, we report the results of training on the AudioSet dataset, which is a million-scale dataset of `in-the-wild' videos.
Note that, this dataset has never been used for training synchronization models,
due to the low computational efficiency in previous models.
For a fair comparison, we also trained the baseline method \cite{sparse2022iashin} on AudioSet. 
Although the audio-visual correspondence in this dataset is not present in all videos, the model was able to benefit from the large amount, yet noisy, training data.
We provide the results of an extensive ablation study in an ArXiv version.

\subsection{Results on Additional Capabilities}
\label{sec:additional_capabilities}

\begin{figure}
  \raggedleft
  \resizebox{0.98\linewidth}{!}{
    \includegraphics{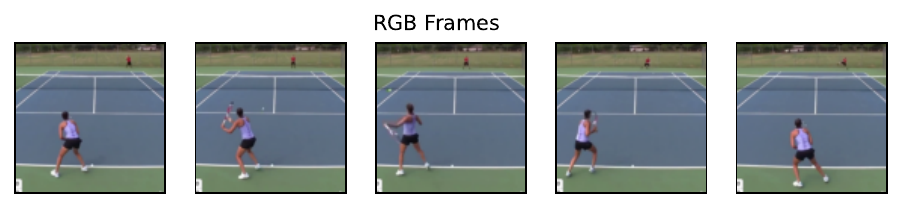}
  }
  \resizebox{1.0\linewidth}{!}{
    \includegraphics{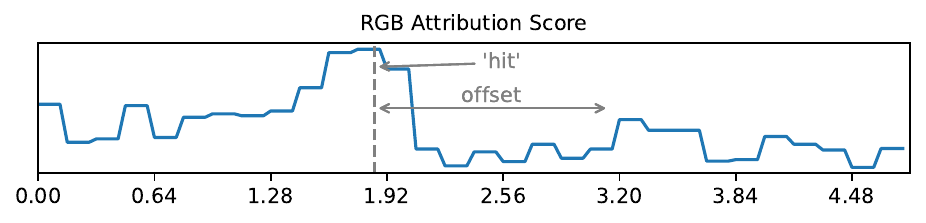}
  }
  \resizebox{1.0\linewidth}{!}{
    \includegraphics{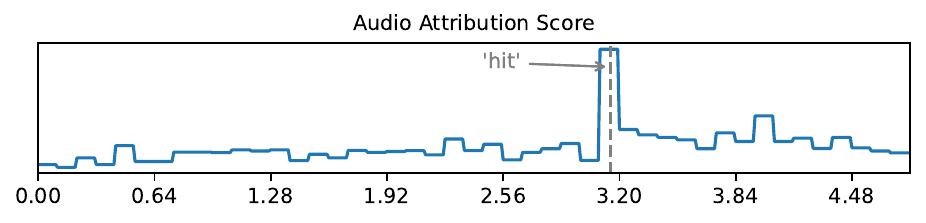}
  }
  \resizebox{1.0\linewidth}{!}{
    \includegraphics{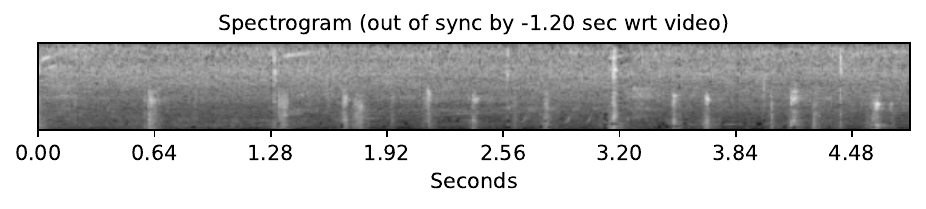}
  }
  \caption{
    \textbf{Visualization of evidence attribution.}
    The moment of `hitting the ball' and the ground truth `offset' are highlighted in both streams.
  }
  \label{fig:evidence_vis}
\end{figure}

\paragraph*{Evidence attribution.}

In Fig.~\ref{fig:evidence_vis}, we visualize the evidence used by the model to make the synchronization predictions for a test video from VGGSound-Sparse as described in §\ref{sec:evidence}.  
The attribution values are min-max scaled. 
We use the \textit{Synchformer} that was trained on the VGGSound dataset.  The video is out-of-sync by $-1.2$ sec, and the model predicts the correct offset.  The model is able to attribute evidence more precisely in the audio stream, i.e. only peaks at $\sim$3.15 sec.
\paragraph*{Predicting syncronizability results.}

\begin{figure}[t]
    \begin{minipage}{0.239\textwidth}
        \centering
        \includegraphics[width=\linewidth]{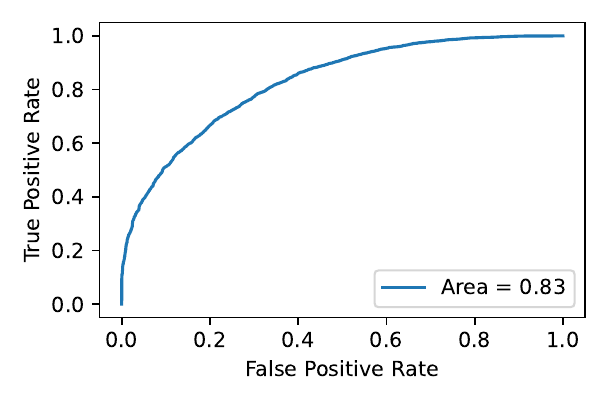}
    \end{minipage}
    \begin{minipage}{0.239\textwidth}
        \centering
        \includegraphics[width=\linewidth]{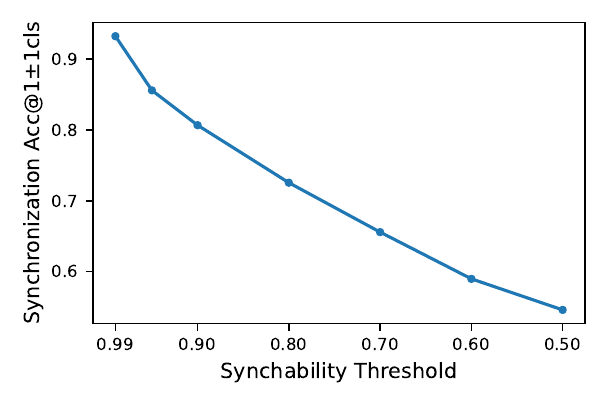}
    \end{minipage}%
    \caption{
      \textbf{Predicting synchronizability with Synchformer.}
      \textit{Left}: ROC curve.
      \textit{Right}: the synchronization performance on videos that were ranked by
      the synchronizability model.
      The results are reported on VGGSound-Sparse.
    }
    \label{tab:syncability}
\end{figure}

In Fig.~\ref{tab:syncability} (left), we use the model that was trained for synchronization,  fine-tune this for synchronizability using AudioSet, and test it on VGGSound-Sparse.
The area under the curve is 0.83, setting the baseline for the task.
In Fig.~\ref{tab:syncability} (right), we display the offset prediction performance for videos from the VGGSound-Sparse dataset, filtered by the synchronizability model based on the confidence of video synchronizability.  This gives an evaluation of the usefulness of synchronizability. 
As anticipated, the offset prediction performance is superior for audio-visual clips that the synchronizability model expresses confidence in.

\section{Conclusion}
\label{sec:conclusion}

We proposed a novel transformer-based model for audio-visual synchronization which
outperforms the state-of-the-art method by a significant margin in both \textit{dense} and
\textit{sparse} settings.
We achieve this by decoupling the training of feature extractors from the synchronization module.
To this end, we introduced a novel \textit{Segment AVCLIP}
pre-training method for segment-level contrastive learning
and a \textit{Synchformer} synchronization module which operates on the `frozen' feature extractors
that promotes adaptability for other downstream tasks, such as synchronizability prediction, which is a
novel task introduced in this paper.
Finally, we explored a method for evidence attribution that highlights the evidence used by the model
to make synchronization predictions.


\bibliographystyle{IEEEbib}
\bibliography{refs}

\begin{thebibliography}{10}

\bibitem{Chung2019perfect}
S.-W. Chung, J.~S. Chung, and H.-G. Kang,
\newblock ``Perfect match: Improved cross-modal embeddings for audio-visual
  synchronisation,''
\newblock in {\em ICASSP}, 2019.

\bibitem{Chung16a}
J.~S. Chung and A.~Zisserman,
\newblock ``Out of time: automated lip sync in the wild,''
\newblock in {\em Workshop on Multi-view Lip-reading, ACCV}, 2016.

\bibitem{Afouras20b}
T.~Afouras, A.~Owens, J.~S. Chung, and A.~Zisserman,
\newblock ``Self-supervised learning of audio-visual objects from video,''
\newblock in {\em ECCV}, 2020.

\bibitem{arandjelovic2018objects}
R.~Arandjelovic and A.~Zisserman,
\newblock ``Objects that sound,''
\newblock in {\em ECCV}, 2018.

\bibitem{Owens18audio-visual}
A.~Owens and A.~Efros,
\newblock ``Audio-visual scene analysis with self-supervised multisensory
  features,''
\newblock in {\em ECCV}, 2018.

\bibitem{chen2021audio}
H.~Chen, W.~Xie, T.~Afouras, A.~Nagrani, A.~Vedaldi, and A.~Zisserman,
\newblock ``Audio-visual synchronisation in the wild,''
\newblock in {\em BMVC}, 2021.

\bibitem{sparse2022iashin}
V.~Iashin, W.~Xie, E.~Rahtu, and A.~Zisserman,
\newblock ``Sparse in space and time: Audio-visual synchronisation with
  trainable selectors,''
\newblock in {\em BMVC}, 2022.

\bibitem{gemmeke2017audio}
J.~Gemmeke, D.~Ellis, D.~Freedman, A.~Jansen, W.~Lawrence, R.~C. Moore,
  M.~Plakal, and M.~Ritter,
\newblock ``Audio set: An ontology and human-labeled dataset for audio
  events,''
\newblock in {\em ICASSP}, 2017.

\bibitem{hershey1999audio}
J.~Hershey and J.~Movellan,
\newblock ``Audio vision: Using audio-visual synchrony to locate sounds,''
\newblock {\em NeurIPS}, 1999.

\bibitem{slaney2000facesync}
M.~Slaney and M.~Covell,
\newblock ``Facesync: A linear operator for measuring synchronization of video
  facial images and audio tracks,''
\newblock {\em NeurIPS}, 2000.

\bibitem{chung2016lip}
J.~S. Chung and A.~Zisserman,
\newblock ``Lip reading in the wild,''
\newblock in {\em ACCV}, 2016.

\bibitem{rabiner1993fundamentals}
L.~Rabiner and B.-H. Juang,
\newblock {\em Fundamentals of speech recognition},
\newblock Prentice-Hall, Inc., 1993.

\bibitem{halperin2019dynamic}
T.~Halperin, A.~Ephrat, and S.~Peleg,
\newblock ``Dynamic temporal alignment of speech to lips,''
\newblock in {\em ICASSP}, 2019.

\bibitem{Khosravan19}
N.~Khosravan, S.~Ardeshir, and R.~Puri,
\newblock ``On attention modules for audio-visual synchronization,''
\newblock in {\em Workshop on Sight and Sound, CVPR}, 2019.

\bibitem{kim2021end}
Y.~J. Kim, H.~S. Heo, S.-W. Chung, and B.-J. Lee,
\newblock ``End-to-end lip synchronisation based on pattern classification,''
\newblock in {\em SLT Workshop}, 2021.

\bibitem{kadandale2022vocalist}
V.~S. Kadandale, J.~F. Montesinos, and G.~Haro,
\newblock ``Vocalist: An audio-visual synchronisation model for lips and
  voices,''
\newblock in {\em Interspeech}, 2022.

\bibitem{gupta2023modeformer}
A.~Gupta, R.~Tripathi, and W.~Jang,
\newblock ``{ModEFormer}: Modality-preserving embedding for audio-video
  synchronization using transformers,''
\newblock in {\em ICASSP}. IEEE, 2023.

\bibitem{Chen20a}
H.~Chen, W.~Xie, A.~Vedaldi, and A.~Zisserman,
\newblock ``{VGG-Sound}: A large-scale audio-visual dataset,''
\newblock in {\em ICASSP}, 2020.

\bibitem{gong2021ast}
Y.~Gong, Y.~Chung, and J.~Glass,
\newblock ``{AST: Audio Spectrogram Transformer},''
\newblock in {\em Interspeech}, 2021.

\bibitem{patrick2021keeping}
P.~Mandela, D.~Campbell, Y.~Asano, I.~Misra, F.~Metze, C.~Feichtenhofer,
  A.~Vedaldi, and J.~F. Henriques,
\newblock ``Keeping your eye on the ball: Trajectory attention in video
  transformers,''
\newblock in {\em NeurIPS}, 2021.

\bibitem{bertasius2021space}
G.~Bertasius, H.~Wang, and L.~Torresani,
\newblock ``Is space-time attention all you need for video understanding?,''
\newblock in {\em ICML}, 2021.

\bibitem{vaswani2017attention}
A.~Vaswani, N.~Shazeer, N.~Parmar, J.~Uszkoreit, L.~Jones, A.~N. Gomez,
  {\L}.~Kaiser, and I.~Polosukhin,
\newblock ``Attention is all you need,''
\newblock in {\em NeurIPS}, 2017.

\bibitem{devlin2018bert}
J.~Devlin, M.-W. Chang, K.~Lee, and K.~Toutanova,
\newblock ``{BERT}: Pre-training of deep bidirectional transformers for
  language understanding,''
\newblock in {\em NAACL: HLT}, 2019.

\bibitem{radford2021learning}
A.~Radford, J.~W. Kim, C.~Hallacy, A.~Ramesh, G.~Goh, S.~Agarwal, et~al.,
\newblock ``Learning transferable visual models from natural language
  supervision,''
\newblock in {\em ICML}, 2021.

\bibitem{oord2018representation}
A.~Oord, Y.~Li, and O.~Vinyals,
\newblock ``Representation learning with contrastive predictive coding,''
\newblock {\em arXiv preprint arXiv:1807.03748}, 2018.

\bibitem{afouras2018lrs3}
T.~Afouras, J.~S. Chung, and A.~Zisserman,
\newblock ``{LRS3-TED}: a large-scale dataset for visual speech recognition,''
\newblock {\em arXiv preprint arXiv:1809.00496}, 2018.

\bibitem{he2016deep}
K.~He, X.~Zhang, S.~Ren, and J.~Sun,
\newblock ``Deep residual learning for image recognition,''
\newblock in {\em CVPR}, 2016.

\bibitem{xie2018rethinking}
S.~Xie, C.~Sun, J.~Huang, Z.~Tu, and K.~Murphy,
\newblock ``Rethinking spatiotemporal feature learning: Speed-accuracy
  trade-offs in video classification,''
\newblock in {\em ECCV}, 2018.

\bibitem{carion2020end}
N.~Carion, F.~Massa, G.~Synnaeve, N.~Usunier, A.~Kirillov, and S.~Zagoruyko,
\newblock ``End-to-end object detection with transformers,''
\newblock in {\em ECCV}, 2020.

\bibitem{goyal2017something}
R.~Goyal, S.~Kahou, V.~Michalski, J.~Materzynska, et~al.,
\newblock ``The "something something" video database for learning and
  evaluating visual common sense,''
\newblock in {\em ICCV}, 2017.

\end{thebibliography}

\appendix
\section{Implementation Details}
\label{sec:supp_implementation}
We use $0.64$ seconds long segments.
The RGB and audio streams are resampled to $25$ fps and $16$ kHz, following best practices suggested
in \cite{sparse2022iashin}.
We extract $128$ mel-spectrogram channels from audio segments of $25$ ms with a hop of $10$ ms.
RGB frames are cropped to $224^2$.
This gives us the input of size ($128 \times 66 \times 1$) for audio and
($16 \times 224 \times 224 \times 3$) for RGB segments ($A_s, V_s$).
After applying the feature extractors,
we obtain audio and visual features ($a_s, v_s$) of size ($12 \times 6 \times 768$) and
($8 \times 14 \times 14 \times 768$), respectively.
During the first stage, we use the batch size $B=2$ and $S=14$ segments per video (=8.96 seconds),
totalling $BS=28$ elements in the contrastive pool.
In our experiments, higher $B$ or using the momentum queue did not translate
to better synchronization performance.
During the second stage, we use $B=16$ and $S=14$ half-overlapped segments per video (=4.8 seconds).
The transformer $\mathcal{T}$ has $3$ layers and $8$ attention heads and $d=768$ hidden units.

Both stages were trained on 16 (32) AMD Instinct MI250 GPUs.
Half precision was used during training.
We note, that the bottleneck for training the synchronization module (Stage II) was associated with 
data I/O rather than GPU, i.e. the training of the Stage II can be conducted on a modest GPU.
The code is publicly available at: \href{https://github.com/v-iashin/Synchformer}{\color{new} \textbf{\texttt{github.com/v-iashin/Synchformer}}}

\begin{table}[t]
  \centering
  \small
  \setlength{\tabcolsep}{5pt}
  \begin{tabular}{c c c c c}
    \toprule
    \textbf{Pre-train}    & \textbf{Pre-train}     & \textbf{Frozen}    & \textbf{Segment} & \textbf{VGS-Sparse}      \\
    \textbf{unimodal}     & \textbf{AVCLIP}        & \textbf{feat extr} & \textbf{overlap} & \textbf{Acc@1 / $\pm$1 cls} \\
    \midrule
    \ding{55} & \ding{55} & \ding{55} & \ding{52} & ~~5.3 / 10.7                   \\
    \ding{52} & \ding{55} & \ding{55} & \ding{52} & ~~8.4 / 22.8                   \\
    \ding{52} & \ding{52} & \ding{55} & \ding{52} & 21.0 / 40.9                   \\
    \ding{52} & \ding{52} & \ding{52} & \ding{55} & 40.1 / 57.0                   \\
    \ding{52} & \ding{52} & \ding{52} & \ding{52} & \textbf{43.7} / \textbf{60.2} \\
    \bottomrule
  \end{tabular}
  \caption{
    \textbf{Ablation study: initialization, training, segment overlap.}
    The results are reported on the test set of VGGSound-Sparse.
    The metrics are top-1 accuracy across 21 offset classes without and with $\pm$1 class tolerance
    (indicated by the value before and after `/').
    `Pre-train unimodal' -- if audio and visual feature extractors are pre-trained on
    AudioSet and Something-Something v.2 \cite{goyal2017something} for audio and action recognition;
    `Pre-train AVCLIP' -- if the feature extractors are pre-trained with Segment AVCLIP (§\ref{sec:training});
    `Frozen feat extr' -- if feature extractors' weights are `frozen' (not updated) during synchronization training; 
    `Segment overlap' -- if the segments are overlapped by half of the segment length during synchronization training.
    The performance of the final model is reported in the last row.
  }
  \label{tab:ablation}
\end{table}

\begin{table}[t]
  \centering
  \small
  \begin{tabular}{r r r c}
        \toprule
        \textbf{Segment}    & \textbf{\# of segments}          & \textbf{Input}     &      \\
        \textbf{length}     & \textbf{(during}              & \textbf{length}    & \textbf{VGS-Sparse}      \\
        \textbf{(seconds)}  & \textbf{2$^\text{nd}$ stage)} & \textbf{(seconds)} & \textbf{Acc@1 / $\pm$1 cls}\\
        \midrule
        2.56          & 3                            & 5.12                      & 26.2 / 49.3                   \\
        1.28          & 7                            & 5.12                      & 40.0 / 61.0                   \\
        {\color{blue} 0.64} & {\color{blue} 14} & {\color{blue} 4.80}            & {\color{blue} 43.7 / 60.2}    \\
        \textbf{0.64} & \textbf{15}                  & \textbf{5.12}             & \textbf{44.7} / \textbf{62.1} \\
        0.32          & 31                           & 5.12                      & 41.2 / 58.2                   \\
        0.16          & 63                           & 5.12                      & 31.2 / 49.7                   \\
        0.08          & 127                          & 5.12                      & 25.9 / 46.0                   \\
        \bottomrule
  \end{tabular}
\caption{\textbf{Ablation study: segment length}.
    The results are reported on the test set of VGGSound-Sparse.
    The metrics are top-1 accuracy across 21 offset classes without and with $\pm$1 class tolerance
    (indicated by the value before and after `/').
    The preferred model is in {\color{blue} blue}. The best model is in \textbf{bold}.
}
\label{tab:segment_length}
\end{table}

\begin{table}[t]
    \centering
    \small
    \setlength{\tabcolsep}{2pt}
    \begin{tabular}{l r c c}
        \toprule
        \textbf{      } & \textbf{Feature}    & \textbf{LRS3-FS       } & \textbf{VGS-Sparse}     \\
        \textbf{Method} & \textbf{extractors} & \textbf{Acc@1 / $\pm$1 cls} & \textbf{Acc@1 / $\pm$1 cls} \\
        \midrule
        SparseS \cite{sparse2022iashin} & ResNet-18 + S3D    & 80.8 / 96.9 & 33.5 / 51.2$^*$ \\
        Ours                            & ResNet-18 + S3D    & 82.6 / 98.9 & 33.9 / 54.7~~ \\
        Ours                            & AST + Mformer      & \textbf{86.5} / \textbf{99.6} & \textbf{43.7} / \textbf{60.2}~~ \\
        \bottomrule
    \end{tabular}
    \caption{\textbf{Ablation study: feature extractors}. 
        The synchronization results are reported on the test sets of LRS3-FS (`Full Scene') and VGGSound-Sparse (VGS-Sparse) datasets.
        The metrics are top-1 accuracy across 21 offset classes without and with $\pm$1 class tolerance
        (indicated by the value before and after `/').
        ResNet-18 and S3D were initially pre-trained for audio recognition on VGGSound and action recognition 
        on Kinetics 400, while AST and Motionformer (Mformer) were pre-trained on AudioSet and Something-Something v.2.
        $^*$~--~the synchronization model was pre-trained on dense signals first (LRS3-FS).
        }
    \label{tab:feat_extractors}
\end{table}

\section{Ablation Studies}
\label{sec:supp_ablation}

\paragraph*{Initialization, training, and segment overlap.}
As shown in Tab.~\ref{tab:ablation}, training the weights of the synchronization module 
from scratch, i.e.\ without pre-training of the feature extractors, significantly reduces the performance.
We also found that initializing feature extractors with pre-trained weights from the unimodal pre-training,
e.g.\ on AudioSet and Something-Something v.2~\cite{goyal2017something} for audio and action recognition,
respectively, improves the performance, yet it took significantly more iterations for
the model to train and it still underperformed compared to other settings.
The initialization with the weights from Segment AVCLIP pre-training (§\ref{sec:training}) allows the model
to start learning faster and reach higher performance.
However, we noticed that if the weights of the feature extractors are kept trainable during the synchronization
training, the model overfits the training data and performs poorly on the test set.
Freezing the feature extractors during the training of the synchronization module allows 
increased batch size and learning rate
that speeds up and stabilizes the training allowing better performance.
Finally, we found that training the model with half-overlapped segments during the synchronization training
improves the performance significantly.

\paragraph*{Segment length.}
We compare the model performance across different segment lengths and report the results in Table~\ref{tab:segment_length}.
For all selected segment lengths, except for 0.64 seconds, we conduct the full training pipeline as used for the final model (see §\ref{sec:training}).
For the experiment with 0.64-second segments, we trained only the synchronization module on longer inputs (15 segments).
Note that we conduct this additional experiment with 0.64-second segments to make the total input length in seconds equal to 5.12
for a fair comparison.
All models for this ablation were trained on VGGSound.
We make two observations: (i) the model trained with 0.64-second segments performs the best (44.7 / 62.1), 
(ii) extending the input length by one segment improves performance as more information is provided to the model to perform
the task (43.7 / 60.2 vs 44.7 / 62.1).

\paragraph*{Feature extractors.}
In Table~\ref{tab:feat_extractors}, we show the performance of our approach with another set of 
audio and visual feature extractors.
In particular, instead of the AST and Motionformer that were used in this work, we employ S3D \cite{xie2018rethinking} and 
ResNet-18 \cite{he2016deep} as in SparseSync \cite{sparse2022iashin}.
By matching feature extractors to~\cite{sparse2022iashin}, we aim to highlight the importance of the training approach
introduced in this work.
We note that even though SparseSync was additionally pre-trained on LRS3 (`Full Scene') for synchronization before training
on VGGSound, our approach yields superior results on both dense and sparse datasets.
Also, our method allows pre-training of feature extractors to be separated from training the synchronization module, which
gives a larger computation budget for feature extractors.
By replacing ResNet-18 and S3D with AST and Motionformer, that have almost one order of magnitude more trainable parameters, 
we improve performance significantly.

\begin{table}[t]
    \small
    \begin{tabular}{c c c}
        \multicolumn{3}{l}{\textbf{Training data: LRS3 (`Full Scene')} $\downarrow$} \\
        \toprule
        \multicolumn{2}{c}{\textbf{Stage}} & \textbf{LRS3 (`Full Scene')}     \\
        \textbf{I} & \textbf{II} & \textbf{($N = 5966 \times 2$)} \\
        \midrule
        i                 &  A                 & 86.4 / 99.6     \\
        ii                &  B                 & 86.6 / 99.6     \\
        \bottomrule
    \end{tabular}
    \\
    \begin{tabular}{c c c c}
        \multicolumn{4}{l}{\textbf{Training data: VGGSound} $\downarrow$} \\
        \toprule
        \multicolumn{2}{c}{\textbf{Stage}} & \textbf{VGGSound-Sparse}       & \textbf{VGGSound-Sparse (Clean)}       \\
        \textbf{I}        & \textbf{II}    & \textbf{($N = 542 \times 25$)} & \textbf{($N = 439 \times 1$)} \\
        \midrule
        iii               &    C           & 43.4 / 60.3                    & 51.8 / 72.1                   \\
        iv                &    D           & 44.1 / 60.2                    & 54.0 / 68.1                   \\
        iv                &    E           & 43.8 / 60.2                    & 55.1 / 71.4                   \\
        \bottomrule
    \end{tabular}
    \\
    \begin{tabular}{c c c c}
        \multicolumn{4}{l}{\textbf{Training data: AudioSet} $\downarrow$} \\
        \toprule
        \multicolumn{2}{c}{\textbf{Stage}} & \textbf{VGGSound-Sparse}       & \textbf{VGGSound-Sparse (Clean)}       \\
        \textbf{I}        & \textbf{II}    & \textbf{($N = 542 \times 25$)} & \textbf{($N = 439 \times 1$)} \\
        \midrule
        v                 &    F           & 46.4 / 66.8                    & 53.8 / 77.9                   \\
        vi                &    G           & 47.2 / 67.4                    & 55.4 / 77.2                   \\
        \bottomrule
    \end{tabular}
    \caption{\textbf{Ablation study: results reproducibility}. 
        The synchronization results are reported on test sets of LRS3 (`Full Scene'), VGGSound-Sparse, 
        and VGGSound-Sparse (Clean).
        The checkpoints in the top table were trained on LRS3 (`Full Scene'), VGGSound was used for the middle table,
        and AudioSet for the bottom table.
        $N$ indicates the size of the dataset multiplied by the number of times offsets were randomly sampled.
        The metrics are top-1 accuracy across 21 offset classes without and with $\pm$1 class tolerance
        (indicated by the value before and after `/').
        The checkpoint IDs are i--vi and A--G for Stages I and II, respectively.
        }
    \label{tab:reproducibility}
\end{table}

\section{Note on Results Reproducibility}
To get robust estimates on the performance of our model, we average performance across multiple \textit{training} runs for
LRS3 (`Full Scene'), VGGSound, and AudioSet and report the mean in the main part of this paper.
The original values of experiments that were used for averaging are reported in Table~\ref{tab:reproducibility}.

In addition, we conduct multiple rounds of offset sampling for each video in the test sets of the evaluation datasets to
increase the robustness of an individual model.
For instance, we sample an offset and a starting point of a temporal crop 2 times ($\times 2$ in the 
Table~\ref{tab:reproducibility}) for LRS3 (`Full Scene') and 25 times ($\times 25$) for VGGSound-Sparse. 
The results across the rounds are averaged.
Notice that we only use one set of offsets for VGGSound-Sparse \textit{(Clean)} considering the difficulty of annotating
it manually.

In our experiments, we found that one could replicate the training dynamics (shapes of the loss curve) from previous runs
during Stage I (feature extractors pre-training) for all datasets.
For Stage II, the training dynamics are predominantly similar on LRS3 (`Full Scene') and AudioSet but vary slightly for 
VGGSound.

We found that one could replicate the performance of the LRS3 model up to 0.1\% even if both stages were re-trained anew.
However, the performance of the VGGSound and AudioSet models vary which results in 
variations of up to 0.8 percent points when evaluated on VGGSound-Sparse.

Finally, we noticed a strong variation (±3\% points) in the performance of the model on the VGGSound-Sparse (Clean) across runs
This is likely due to its size (439 videos) and the fact that only one offset per video is evaluated.

The variation in performance across runs comes from multiple sources that include (but not limited to):
\begin{enumerate}
    \item Variation in training during the first stage (segment-level contrastive pre-training);
    \item Variation in training during the second stage (synchronization module training), e.g.\ due to 
    unique offset sampling at every iteration, which we call `path dependency', and accumulated variation from the first stage;
    \item Variation in the evaluation due to random offset sampling (e.g.\ for every video in VGGSound-Sparse, 
    we randomly sample 25 offsets and average the performance across them as described above).
\end{enumerate}

We tested if the variation is caused by the weights of Stage I.
To this end, we trained a synchronization model (Stage II, iv--E) with the feature extractors that were initialized with the 
same weights as the other run of Stage II (iv--D). 
However, the results didn't show any difference in variation which suggests that variation is not caused by the 
the difference is the weights of Stage I alone.

\end{document}